\documentclass{article}
\usepackage{arxiv}
\usepackage[utf8]{inputenc} 
\usepackage[T1]{fontenc}    
\usepackage{hyperref}       
\usepackage{url}            
\usepackage{booktabs}       
\usepackage{amsfonts}       
\usepackage{nicefrac}       
\usepackage{microtype}      
\usepackage{lipsum}		
\usepackage{graphicx}
\usepackage{natbib}
\usepackage{doi}
\usepackage{amsmath}
\usepackage{amsthm}

\usepackage{pgfplots}
\pgfplotsset{compat=1.17}
\usepackage{xspace}
\usepackage{amsfonts}
\usepackage{algorithm}
\usepackage{algorithmic}
\usepackage{multirow}
\usepackage{float}
\usepackage{colortbl}

\AtBeginDocument{%
  \providecommand\BibTeX{{%
    \normalfont B\kern-0.5em{\scshape i\kern-0.25em b}\kern-0.8em\TeX}}}

\begin{document}

\title{Multi-Objective Evolutionary Algorithms with Sliding Window Selection for the Dynamic Chance-Constrained Knapsack Problem}

\author{
        {Kokila Kasuni Perera} \\
        Optimisation and Logistics\\
	School of Computer and Mathematical Sciences\\
        The University of Adelaide\\
        Adelaide, Australia \\
        \And
	{Aneta Neumann} \\
 Optimisation and Logistics\\
	School of Computer and Mathematical Sciences\\
        The University of Adelaide\\
        Adelaide, Australia \\
}

\date{}

\maketitle

\sloppy

\begin{abstract}
Evolutionary algorithms are particularly effective for optimisation problems with dynamic and stochastic components. We propose multi-objective evolutionary approaches for the knapsack problem with stochastic profits under static and dynamic weight constraints. The chance-constrained problem model allows us to effectively capture the stochastic profits and associate a confidence level to the solutions' profits. We consider a bi-objective formulation that maximises expected profit and minimises variance, which allows optimising the problem independent of a specific confidence level on the profit. We derive a three-objective formulation by relaxing the weight constraint into an additional objective. We consider the GSEMO algorithm with standard and a sliding window-based parent selection to evaluate the objective formulations. Moreover, we modify fitness formulations and algorithms for the dynamic problem variant to store some infeasible solutions to cater to future changes. We conduct experimental investigations on both problems using the proposed problem formulations and algorithms. Our results show that three-objective approaches outperform approaches that use bi-objective formulations, and they further improve when GSEMO uses sliding window selection.
\end{abstract}


\maketitle

\section{Introduction}
The dynamic and stochastic nature of real-world optimisation problems makes it challenging to find solutions. The optimisation methods that perform significantly for the static and deterministic problems may perform differently in addressing dynamic and stochastic problems \cite{surveryConstraints2005}. Designing the methods to capture the dynamic and stochastic problem parameters is essential to obtain more reliable solutions \cite{Neumann2020}.

Dynamic optimisation problems have components that effectively change the objective function, a constraint or a decision variable over time \cite{NGUYEN20121}. A dynamic change to one problem component may also affect other components, such as a change in a constraint, which may change the feasibility of solutions or move the optima. The frequency and magnitude of change can characterise the dynamic component of the problem. 
In this work, we focus on the dynamic constraint bound, which changes in regular time intervals ($\tau$), and the value increases or decreases with a random value. The magnitude of change ($\gamma$) takes off a particular distribution. These characteristics provide a better understanding of the effects of the dynamic changes on the problem. The optimisation methods for dynamic problems require identifying the optima and also tracking the moving optima  \cite{NGUYEN20121,DBLP:conf/gecco/BossekN023}. 

The uncertainties in the problem parameters or environment are another challenge. They can lead to inaccuracies in estimating decision variables or evaluating constraints and objective functions \cite{surveryConstraints2005}. Furthermore, this can cause optimisation methods to produce a suboptimal or infeasible solution if they poorly capture the uncertain parameters. The chance constraint has shown success in addressing such stochastic problems \cite{DBLP:conf/ppsn/NeumannN20, submodularccAAAI20,Liu_EA_for_CC,TCHTTP,ERDCCCMCP,POCCMSP}. The chance constraint defines a small probability value $\alpha<0.5$ as an upper bound on the probability that the constraint is violated. The chance constraint indicates that we tolerate the constraint violation by a solution if that probability is small. The chance-constrained problem models allow us to associate the reliability of a solution \cite{Deb2009RBO} and to design practical applications for stochastic optimisation problems in many fields, such as logistics \cite{cc-schedulingAAAI2015} and mining \cite{mining_2021,yue_stockpile_2021}.

This work focuses on two variations of the classical knapsack problems: stochastic and dynamic constraints. The classical knapsack problem has deterministic problem parameters and is an NP-hard combinatorial optimisation problem. Given a set of elements with deterministic profits and weights, the knapsack problem aims to find a packing plan that maximises the total profit subject to a given weight bound. 
Literature has considered variations of this problem with stochastic weights or stochastic profits \cite{yue_gecco_19, yue_gecco_20, Aneta_ppsn_22,E23OMOEA} as well as dynamic settings \cite{Hirad_ecai_20,10.1007/978-3-319-99253-2_13,U3OEADCHKP,DBLP:journals/ai/RoostapourNNF22,DBLP:journals/tcs/RoostapourNN22}. This work considers the knapsack problem with stochastic profits under static and dynamic weight-bound settings. It is the first study exploring a problem scenario with both stochastic and dynamic constraints. We model the problem as a chance-constrained problem to capture the stochastic profits. The profit-chance constrained knapsack problem model is beneficial in real-world problem scenarios when the confidence levels of the available solutions allow the decision makers more insights in choosing solutions. 
Moreover, the knapsack problem with the dynamic weight constraints represents real-world scenarios where problem constraints change over time due to environmental influences or the parameters in the problem itself.

Evolutionary algorithms have successfully addressed problems with stochastic and dynamic constraints in literature \cite{Neumann2020, DBLP:conf/ppsn/SinghB22, Doer_Neumann_survey_21}. We propose multi-objective evolutionary algorithms for solving the selected problems. The multi-objective evolutionary algorithms use two or more objectives and produce a population of solutions that gives a trade-off of the defined objectives. In particular, multi-objective evolutionary algorithms have been of higher interest as the outcome of these algorithms provides trade-offs of different objectives, which enable the stakeholders to consider alternative solutions. 
\subsection{Related work}
Early works on evolutionary computation for chance-constrained problems consider computationally expensive methods like simulation and sampling to cater for chance constraints \cite{He_Shao_ICIECS_2009, Loughlin_Ranjithan_gecco99, Liu_EA_for_CC, Masutomi_2013_jaciii}. Recent literature shows increased interest in using tail-bound inequalities to deal with chance constraints efficiently \cite{DBLP:conf/ppsn/NeumannN20,Aneta_ppsn_22, yue_gecco_19, Hirad_ecai_20}. Moreover, problems with chance and dynamic constraints have been considered scarcely in the evolutionary computing literature. 

Runtime analysis is essential in studying problems with stochastic and dynamic constraints. The first run time analysis on chance constraint problems analyses (1+1) EA for different cases of the knapsack problem with stochastic weights \cite{neumann_sutton_foga_19}. \cite{yue_runtime_gecco_21} analyse the simple evolutionary algorithms for chance-constrained knapsack problems with uniform weights. The runtime studies \cite{Neumann_ijcai2022} and \cite{CCMakespan_ppsn22} analyse the performance of simple evolutionary methods on different chance-constrained problems. \cite{Neumann_ijcai2022} analyses single- and multi-objective evolutionary approaches for chance-constrained problems with normally distributed random problem variables and proposes evolutionary approaches for chance-constrained minimum spanning tree problems \cite{Neumann_ijcai2022}. In \cite{CCMakespan_ppsn22}, they analyse the run time of random local search and (1+1) EA for the chance-constrained makespan problem. The knapsack problem with a chance constraint on the weights is explored in several recent papers \cite{yue_gecco_19, yue_gecco_20}. These works use well-known deviation inequalities: Chebyshev's inequality and Chernoff bound to approximate the probability of constraint violation. While both papers propose single- and multi-objective evolutionary approaches, \cite{yue_gecco_20} introduces problem-specific operators for the problem. The first runtime analysis of evolutionary multi-objective algorithms for the optimisation of submodular functions with chance constraints has been carried out in \cite{DBLP:conf/ppsn/NeumannN20}. 

Assmi et al. 2020 \cite{Hirad_ecai_20} consider the dynamic chance-constrained knapsack with stochastic weights, which closely aligns with the problem in the current paper. They propose a bi-objective model based on the profit of the solution as one objective and the minimal capacity bound under a given chance constraint setting as the second objective. Additionally, the proposed methods store not only the solutions that adhere to the current weight bound (B) but also infeasible solutions within the vicinity of B, allowing the population to adapt to dynamic changes of B. This later technique has succeeded in \cite{10.1007/978-3-319-99253-2_13}, which studies the knapsack problem with a dynamic weight constraint.

The knapsack problem with stochastic profits is relatively new in the chance-constrained literature. The first paper to study this problem, \cite{Aneta_ppsn_22}, proposes single objective evolutionary approaches using objectives defined using Chebyshev's inequality and Hoeffding bounds to approximate profit. Later, \cite{perera_gecco_23} considers the same problem and proposes a multi-objective formulation for this problem that computes trade-offs concerning the expected value and the variance of solutions motivated by the theoretical analysis presented in \cite{Neumann_ijcai2022}.
\subsection{Our Contribution}
This study addresses two variants of the knapsack problem with stochastic profits under static and dynamic weight constraints. We model these problems as chance-constrained problems and define the dynamic configurations of the weight constraint. According to our knowledge, this is the first study to consider the variation of the knapsack problem with the dynamic weight and profit-chance constraints.

We introduce two objective formulations for these problems. We use a bi-objective fitness formulation initially proposed in \cite{perera_gecco_23} for the profit-chance constrained knapsack problem. This formulation defines objectives based on the expected value and variance of profits. It allows the evolutionary algorithms to optimise the problem independent of the probability value attached to the chance constraint.
Next, we propose a three-objective fitness formulation for the problem motivated by the recent study, which shows the benefits of using three-objective formulations for chance-constrained problems \cite{3objGECCO23}. Literature on chance-constrained knapsack problems is limited to single and bi-objective optimisation. The current work is the first study to propose a three-objective fitness formulation for the profit chance-constrained knapsack problem.

We consider Global Standard Evolutionary Multi-objective Optimiser (GSEMO) with the bi- and three-objective formulations. In addition to the standard parent selection operation, we consider GSEMO with the sliding window technique proposed in \cite{neumann2023fast}, which is particularly effective when dealing with large populations. Moreover, to address the problem under dynamic weight constraint, we introduce variation to GSEMO and modify the fitness formulations to deal with dynamic settings effectively.

We conduct extensive experimental investigations on static and dynamic cases of the profit-chance constrained problem using different problem formulations and evolutionary methods described below. We discuss the performance of the proposed methods using their results for problems under selected confidence levels ($\alpha$). We report the results regarding the maximum profit given by the final population and also an offline error metric indicating how the internal population changes over time.

The paper is structured as follows. Section \ref{sec:prob_static} and \ref{sec:prob_dyn} introduce the knapsack problem with stochastic profits subject to static and dynamic weight constraints and present objective formulations for them. Section \ref{sec:algo} presents the algorithms and techniques we utilise for the target problems. Design of the experimental investigations and analysis of results are given in Section \ref{sec:setups}
and \ref{sec:results}, respectively, followed by the concluding remarks in Section \ref{sec:conclusions}.

\section{Knapsack problem with stochastic profits} \label{sec:prob_static}

First, we consider the profit-chance-constrained knapsack problem subject to a static weight-bound constraint. In this section, we present the formal definition of the problem and introduce the fitness evaluations we consider for the problem.

\subsection{Problem definition \label{sec:prob_def_static}}
This problem is a variation of the classical knapsack problem, which we define as follows. Let the knapsack have $n$ elements with weight $w_i$ and profit $p_i$ for $i=1, \ldots, n$ and weight bound $B$. A solution to the problem is a selection of elements, which we represent as a bit string of length n $x \in \{0,1\}^n$. The weight and profit of the solution $x$ are given by $w(x)=\sum_{i=1}^n w_i\cdot x_i$ and $p(x)=\sum_{i=1}^n p_i\cdot x_i$. 
In the classical knapsack problem, the optimisation goal is to find the best solution $x^*$, which maximises $p(x^*)$ subject to $w(x^*)\leq B$.

When the profits are stochastic, we introduce a chance constraint to the problem to limit the probability that the profits drop below the maximal profit value. Given that the $P$ is the maximal profit a solution can have under the chance constraint with probability $\alpha$, we define the problem as follows:
\begin{eqnarray}
    \label{eq:prob_def}
    & \max P  \\
    \label{eq:profit_cons}
    \mathit {subject~to}   & Pr(p(x) < P) \leq \alpha\\ 
    \label{eq:weigt_cons}
    \mathit {and} & w(x) \leq B
\end{eqnarray}

The goal of solving this problem is to identify a solution that gives the highest profit bound P with the confidence level $\alpha$. In literature, tail-bound inequalities are used to derive the maximal profit of a particular solution under the chance constraint on profit \cite{yue_gecco_19, yue_gecco_20, Aneta_ppsn_22}. In the previous study on the knapsack problem with stochastic profits problem \cite{Aneta_ppsn_22}, the authors propose two profit estimates for the problem based on Chebyshev inequality and Hoefding bound. 

Let $\mu(x)$ be the expected profit of solution $x$ and $v(x)$ be its variance. Then, we derive the maximal profit of $x$ from  Chebyshev inequality~\cite{Book_prob2014} as $P_{Cheb}(x,\alpha)$. Equation \ref{eq:profit_cheby} defines $P_{Cheb}(x,\alpha)$, which is the maximal profit solution can have given that the chance to profit drops below this value is at most $\alpha$.
\begin{eqnarray}
    \label{eq:profit_cheby}
    P_{Cheb}(x, \alpha) = \mu(x) - \sqrt{(1-\alpha)/ \alpha} \cdot \sqrt{v(x)}
\end{eqnarray}

Similarly, we derive the maximal profit for $x$ from Hoefding bound ~\cite{Book_prob2014} as $P_{Hoef}(x,\alpha)$ if the profits are independent and distributed uniformly with the same dispersion. Let the profits of elements be uniformly distributed as $p_i \in [\mu_i-\delta,\mu_i+\delta]$ where $\mu_i$ is the expected profit of element $i$ and $\delta$ is the dispersion of profits. Then, the maximum profit estimate of $x$ subject to chance constraint with a given $\alpha$ is defined as follows:
\begin{eqnarray}
    \label{eq:profit_hoef}
    P_{Hoef}(x,\alpha) = \mu(x) - \delta \cdot \sqrt{\ln(1/\alpha) \cdot 2 \cdot|x|_1}
\end{eqnarray}

\subsection{Objective formulations}
We introduce bi- and three-objective formulations for this problem. We use the bi-objective model proposed in \cite{perera_gecco_23}, which defines objectives based on the expected profit and variance, and for infeasible solutions, the objectives are penalised. Let $v_{max}=\sum_{i=1}^n \sigma^2_i$ be the maximum variance a solution may have for a particular problem instance; we define the bi-objective fitness function ~$f_{2D-static}(x)=(\mu_{st}(x),v_{st}(x))$ as follows:
\begin{equation}
    \label{eq:fitness-1-obj1-static}
    \mu_{st}(x) = \left\{
    \begin{array}{lcl}
        \sum_{i=1}^n {\mu_i x_i} & & {w(x) \leq B}\\
        B - w(x) & & {otherwise}
    \end{array} \right.
\end{equation}
\begin{equation}
    \label{eq:fitness-1-obj2-static}
    v_{st}(x) = \left\{
    \begin{array}{lcl}
        \sum_{i=1}^n {\sigma_i^2x_i} && {w(x) \leq B}\\
        v_{max} && {otherwise}
    \end{array} \right.
\end{equation}

$\mu_{st}$ is a maximising and $v_{st}$ is a minimising objective. Given two solutions, $x$ and $y$, we say $x$ (weakly) dominates $y$ ($x \succeq y$) if $$\mu_{st}(x)\geq \mu_{st}(y) \text{ and } v_{st}(x) \leq v_{st}(y),$$ 
and $x$ strongly dominates $y$ ($x \succ y$) if 
$$x \succeq y \text{ and } (\mu_{st}(x) >\mu_{st}(y) \text{ or } v_{st}(x) < v_{st}(y)).$$
Next, we propose a three-objective fitness evaluation for the same problem ~$f_{3D-static}(x)=(\mu_{st}(x),v_{st}(x),w_{st}(x))$. The first two objectives are the same as in the $f_{2D-static}(x)$. Let $w_{max}=\sum_{i=1}^n w_i$ be the total weight of all elements. We introduce a third objective $w_{st}(x)$ for the formulation by relaxing the weight constraint as follows:
\begin{equation}
    \label{eq:fitness-1-obj3-static}
    w_{st}(x) = \left\{
    \begin{array}{lcl}
        \sum_{i=1}^n {w_i x_i} && {w(x) \leq B}\\
        w_{max} && {otherwise}
    \end{array} \right.
\end{equation}

The new objective is a minimising objective. According to the three-objective model, the solutions $x$ and $y$ holds $x \succeq y$ if 
$$\mu_{st}(x)\geq \mu_{st}(y) \text{ and } v_{st}(x) \leq v_{st}(y) \text{ and } w_{st}(x) \leq w_{st}(y),$$
and holds $x \succ y$ if $x \succeq y$ and 
$$(\mu_{st}(x) >\mu_{st}(y)  \text{ or }   v_{st}(x) < v_{st}(y) \text{ or } w_{st}(x) < w_{st}(y)).$$

\section{Knapsack problem with stochastic profits and dynamic weight bound}\label{sec:prob_dyn}
This section introduces the knapsack problem with stochastic profits subject to a dynamic weight bound. We discuss how the problem definition differs from that under a static bound and introduce objective formulations for the new problem. 
\subsection{Problem definition}
In this problem, the weight bound changes dynamically during the optimisation, but other problem parameters remain the same as defined in Section \ref{sec:prob_def_static}. We consider the \emph{time} during the optimisation in terms of the number of fitness evaluations. We denote the weight bound at a given time $t$ as $B_t$; thus, the weight constraint in the dynamic setting changes as follows:
\begin{equation}
    \label{eq:weigt_cons_dynamic}
    w(x) \leq B_t
\end{equation}

We define the dynamically changing weight bound as follows. Let $B_0$ denote the initial weight bound of a particular problem instance, $\tau$ the time interval between two consecutive changes, and $\gamma_t$ the magnitude of change at time $t$ which is uniformly randomly chosen from $[-\gamma, \gamma]$. Then, we define the weight bound $B_t$ at time $t$ as follows:
\begin{equation}
    \label{eq:dyn_bound}
    B_t = \left\{
    \begin{array}{lcl}
        B_0  & & t=0\\
        B_{(t-1)} + \gamma_t && t\mod\tau = 0 \\
        B_{(t-1)} & & \text{otherwise}
    \end{array} 
    \right.
\end{equation}
\begin{figure}[!t]
    \centering
    \small
    \begin{tabular}{c}
        \includegraphics[scale=0.75]{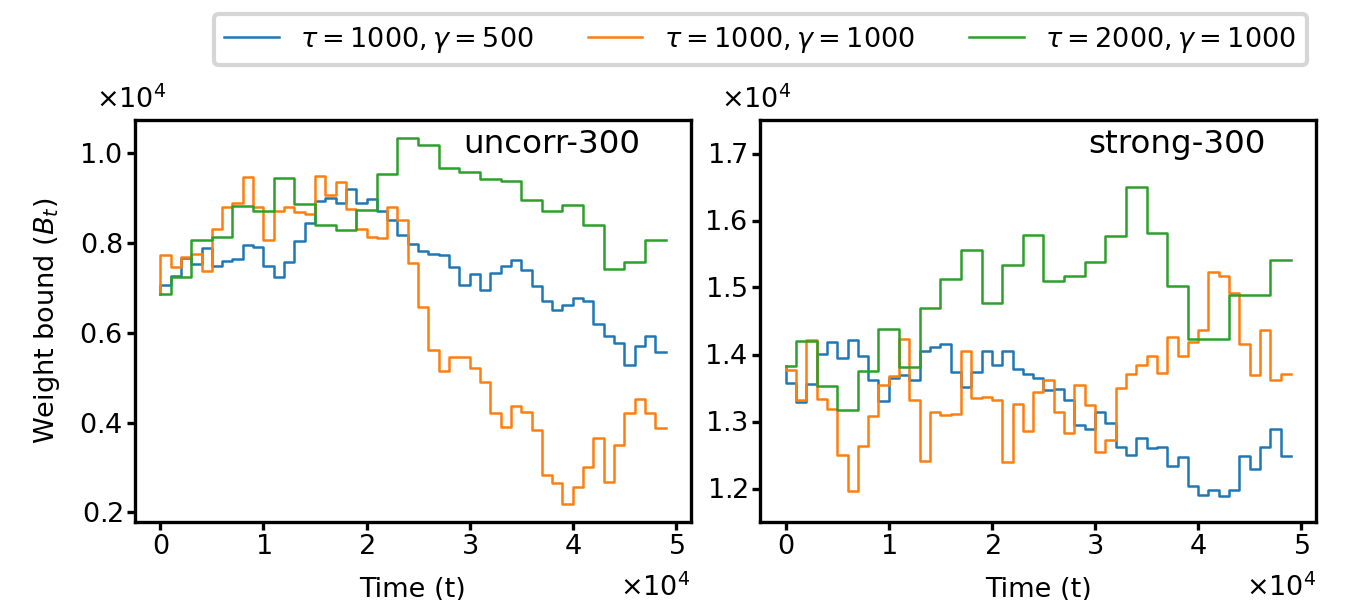}
    \end{tabular}
    \caption{Changes of weight bound over first 50\,000 evaluations for uncorr-300 and strong-300 problem instances}
    \label{fig:dyn_bound}
\end{figure}
Figure \ref{fig:dyn_bound} shows an example of how the weight bound changes over time under different dynamic settings for two problem instances that we use in experimental investigations. The dynamic weight bound does not affect the evaluation of the chance constraint on profit. Therefore, we can employ the same profit estimates ${p}_{Cheb}$ and ${p}_{Hoef}$ defined in Equation \ref{eq:profit_cheby} and \ref{eq:profit_hoef}, respectively. 
\subsection{The objective formulations for dynamic weight bounds \label{sec:fitness-dynamic}}
An infeasible solution in the dynamic setting can become feasible if the weight bound increases later. These solutions can be important to effectively repair the internal populations of an evolutionary approach when dynamic changes occur. In order to accommodate this, we change the objective formulations to not penalise all infeasible solutions. If it is possible for a solution to be feasible at the next weight bound $B_{t+1}$, we allow the algorithms to consider these solutions. Given that $\gamma$ is the maximum magnitude of change, if a solution $x$ holds $w(x) \leq B_t + \gamma$, it can become feasible at the next weight bound $B_t$.
We modify the condition for which the solutions that are penalised in fitness formulations in Equation \ref{eq:fitness-1-obj1-static} and \ref{eq:fitness-1-obj2-static} and introduce the following objective formulations. 

Let $v_{max}=\sum_{i=1}^n \sigma^2_i$ be the maximal variance of a solution to the problem. We define the bi-objective fitness function $f_{2D-dyn}(x)=(\mu_{dyn}(x),v_{dyn}(x))$ for the dynamic bound settings as follows:
\begin{equation}
    \label{eq:fitness-2-obj1}
    \mu_{dyn}(x) = \left\{
    \begin{array}{lcl}
        \sum_{i=1}^n {\mu_i x_i} & & {w(x) \leq B_t + \gamma}\\
        |B_t - w(x)| & & {otherwise}
    \end{array} \right.
\end{equation}
\begin{equation}
    \label{eq:fitness-2-obj2}
    v_{dyn}(x) = \left\{
    \begin{array}{lcl}
        \sum_{i=1}^n {\sigma_i^2x_i} && {w(x) \leq B_t + \gamma}\\
        v_{max} && {otherwise}
    \end{array} \right.
\end{equation}

Similarly, we propose the three-objective fitness function $f_{3D-dyn}(x)=(\mu_{dyn}(x),v_{dyn}(x),w_{dyn}(x))$ for the dynamic settings. The first two objectives have the same definition as above (Equation \ref{eq:fitness-2-obj1} and \ref{eq:fitness-2-obj2}), and the third objective is defined as follows,
\begin{equation}
    \label{eq:fitness-1-obj3}
    w_{dyn}(x) = \left\{
    \begin{array}{lcl}
        \sum_{i=1}^n {w_i x_i} && {w(x) \leq B_t + \gamma}\\
        w_{max} && {otherwise}
    \end{array} \right.
\end{equation}

This new objective $w_{dyn}(x)$ is based on the solution's weight, and the infeasible solutions are penalised accordingly. The dominance between two solutions with respect to the new objective functions $f_{2D-dyn}$ and $f_{3D-dyn}$ is evaluated similarly to how dominance is evaluated when considering objectives of $f_{2D-static}$ and $f_{3D-static}$ following the chronological order of objectives.

\section{Algorithms \label{sec:algo}}

We use multi-objective evolutionary methods to evaluate the performance of the proposed fitness formulations for the target problems. In this section, we present the algorithms that we consider in this paper. 

The first algorithm we consider is the GSEMO algorithm. It is the simplest form of multi-objective evolutionary algorithm and is a good baseline for comparing the performance of different techniques used in evolutionary methods. The algorithmic steps of this method are given in Algorithm \ref{alg:GSEMO}. In this method,  the bit-flip mutation creates an offspring solution from a parent solution selected uniformly randomly from the existing population. If no existing solutions dominate the new solution, it is added to the population, replacing all the solutions it dominates. In this manner, GSEMO maintains a set of non-dominating solutions and optimises them until it meets the stop criterion. In this work, we define the stop criterion in terms of maximum fitness evaluations such that each algorithm run will consider the same number of fitness evaluations ($t_{max}$).

\begin{algorithm}[!t]
    \caption{GSEMO}
    \label{alg:GSEMO}
    \begin{algorithmic}[1]
        \STATE $x \leftarrow \{0\}^n $;
        \STATE $S\leftarrow \{x\}$;
        \STATE $t \leftarrow 1$
        \WHILE{$t < t_{max}$}
            \STATE choose a parent solution $x$ from $S$
            \STATE $y\leftarrow$ flip each bit of $x$ independently with probability of $\frac{1}{n}$;
            \IF{($\not\exists w \in S: w \succ y$)}
                \STATE $S \leftarrow (S \cup \{y\})\backslash \{z\in S \mid y \succeq z\}$ ;
            \ENDIF
            \STATE $t \leftarrow t+1$
        \ENDWHILE
    \end{algorithmic}
\end{algorithm}

We have proposed two fitness formulations for the target problems that use two and three objectives. As the number of objectives grows, we expect to have more trade-offs. While having more trade-offs allows us to get more insights into the population, it may negatively impact the optimisation process by delaying the convergence. The recent literature \cite{neumann2023fast} proposes a sliding window-based selection as an effective technique to eliminate the negative influence of population size and obtain fast Pareto optimisation with multi-objective evolutionary methods. We adopt the sliding window technique from \cite{neumann2023fast} to see its effects on removing negative influences of population size when optimising dynamic and chance-constrained problems.

The sliding window-based parent selection replaces the standard selection method in GSEMO, performed at step 5 in Algorithm \ref{alg:GSEMO}. The steps of the sliding window selection are given in Algorithm \ref{alg:sliding-selection}. Given a fixed length $len_{sw}$, the sliding window is calculated based on the weight bound $B$, the current number of fitness evaluations $t$ and $t_{max}$. The sliding window filters solutions if their weights fit the window. If the filtered subset is non-empty, the parent solution is selected uniformly at random, considering the solutions in this subset or the whole population.

\begin{algorithm}[!t]
\caption{Sliding window selection}
    \begin{algorithmic}[1]
        \REQUIRE $P$, $B$ and $t$
        \IF{$t<t_{max}$}
            \STATE $\hat{B} \leftarrow (t/t_{max})\cdot B$;
            \STATE $\hat{P}=\{x\in P |  \lfloor \hat{B} \rfloor \leq w(x) \leq \lceil \hat{B} + len_{sw} \rceil \}$;
            \IF{$\hat{P}=\emptyset$}
                \STATE $\hat{P} \leftarrow P$;
            \ENDIF
        \ELSE
            \STATE $\hat{P} \leftarrow P$;
        \ENDIF
        \STATE Choose $x$ uniformly at random from $\hat{P}$.
    \end{algorithmic}
\label{alg:sliding-selection}
\end{algorithm}

This sliding window parent selection requires the initial solution of the algorithm to be an empty solution where none of the knapsack elements are selected. For fairness, in all GSEMO variants in this paper, we set the initial solution as empty.
\begin{algorithm}[!t]
\caption{Modified GSEMO for dynamic weight bound}
    \begin{algorithmic}[1]
        \STATE $x \leftarrow \{0\}^n $;
        \STATE $t \leftarrow 1$;
        \IF{$w(x) \leq B_t$}
            \STATE $S_1\leftarrow \{x\}$;
        \ELSE
            \STATE $S_2\leftarrow \{x\}$;
        \ENDIF
        \WHILE{$t < t_{max}$}
            \STATE choose parent solution $x$ from $S_1 \cup S_2$
            \STATE $y\leftarrow$ flip each bit of $x$ independently with probability of $\frac{1}{n}$;
            \STATE $t \leftarrow t+1$
            \IF{$w(y) \leq B_t$ and ($\not\exists w \in S_1: w \succ y$)}
                \STATE $S_1 \leftarrow (S_1 \cup \{y\})\backslash \{z\in S_1 \mid y \succeq z\}$ ;
            \ENDIF
            \IF{$w(y) > B_t$ and ($\not\exists w \in S_2: w \succ y$)}
                \STATE $S_2 \leftarrow (S_2 \cup \{y\})\backslash \{z\in S_2 \mid y \succeq z\}$ ;
            \ENDIF
        \ENDWHILE
    \end{algorithmic}
\label{alg:GSEMO_dyn}
\end{algorithm}

In the dynamic setting, when the bound changes, GSEMO must repair its population by re-evaluating the objective values and adjusting the population to maintain a non-dominated set of solutions. Here, GSEMO needs to consider certain infeasible solutions as they can be helpful in repairing the population when the weight limit changes. The fitness functions $f_{2D-dyn}$ and $f_{3D-dyn}$ facilitate this, as they do not penalise the solutions in the weight range $(B_t, B_t+\gamma]$. However, considering these infeasible solutions in the same population as the feasible solutions can reduce the quality of GSEMO's population. Therefore, we introduce a new population to maintain these specific infeasible solutions separately. The steps of this modified GSEMO are given in Algorithm \ref{alg:GSEMO_dyn}. $S_1$ is the primary population, and $S2$ is the new population for infeasible solutions, and both maintain a non-dominated set of solutions. $S_2$ population that acts as a backlog for infeasible solutions that can become feasible and help repair population $S_1$ when the weight bound changes.

We consider this modified GSEMO with the standard and sliding window-based parent selection for the dynamic settings. The parent selection operation is performed at step 9 of Algorithm \ref{alg:GSEMO_dyn}. The only change to the sliding window selection operation is that it now considers the union of two populations ($S_1\cup S_2$) as $P$, and all steps remain unchanged (see Algorithm \ref{alg:sliding-selection}).
\begin{figure*}[!t]
    \centering
    \small
    \begin{tabular}{c}
         \includegraphics[scale=0.65]{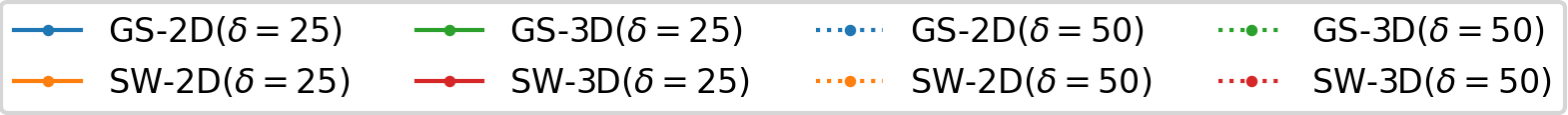}\\
         \includegraphics[scale=0.75]{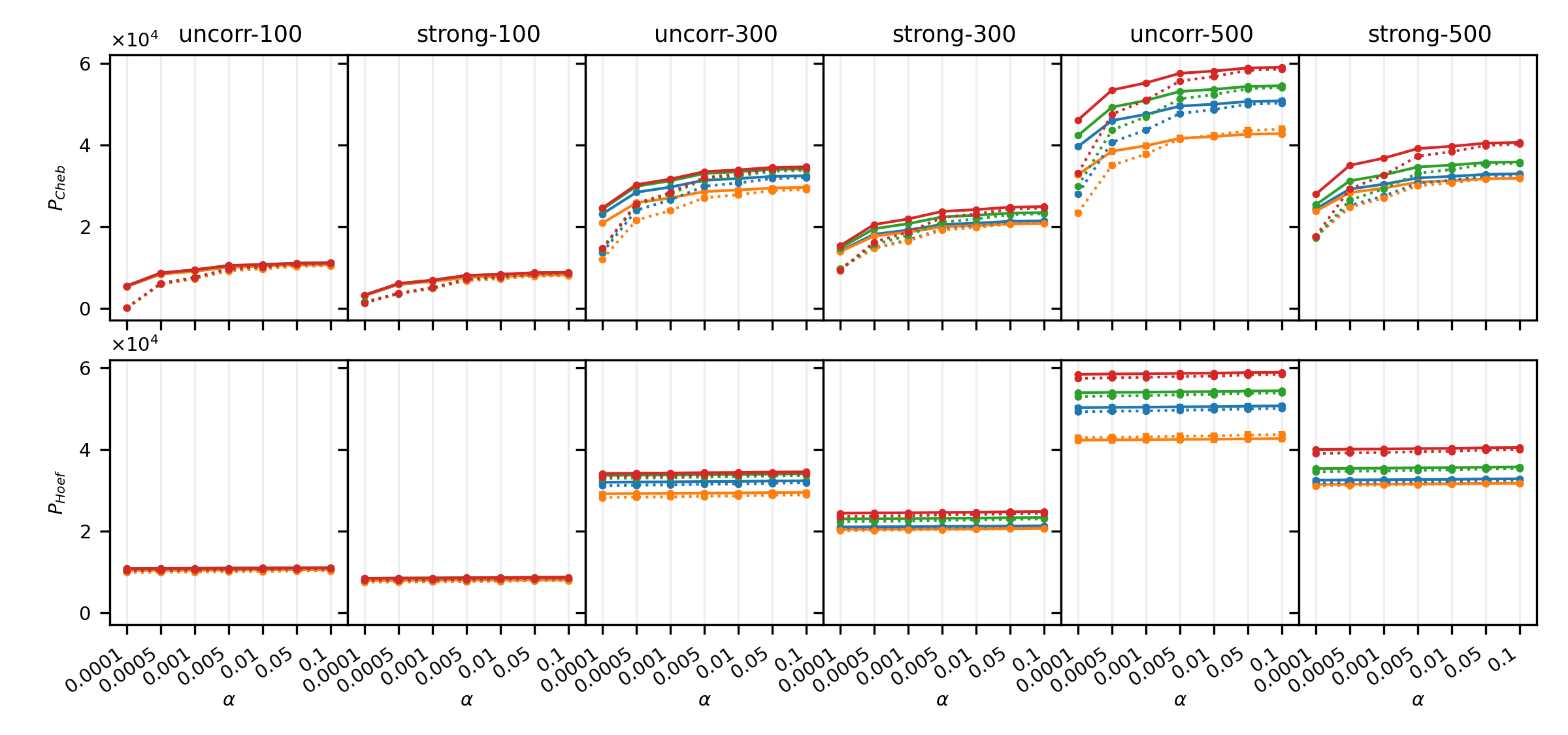}
    \end{tabular}
    \caption{Profit values for static bound setting}
    \label{fig:static_results}
\end{figure*}

\begin{figure}[!t]
    \centering
    \small
    \begin{tabular}{c}
         \includegraphics[scale=0.8]{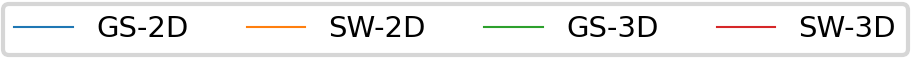}\\
         \includegraphics[scale=0.8]{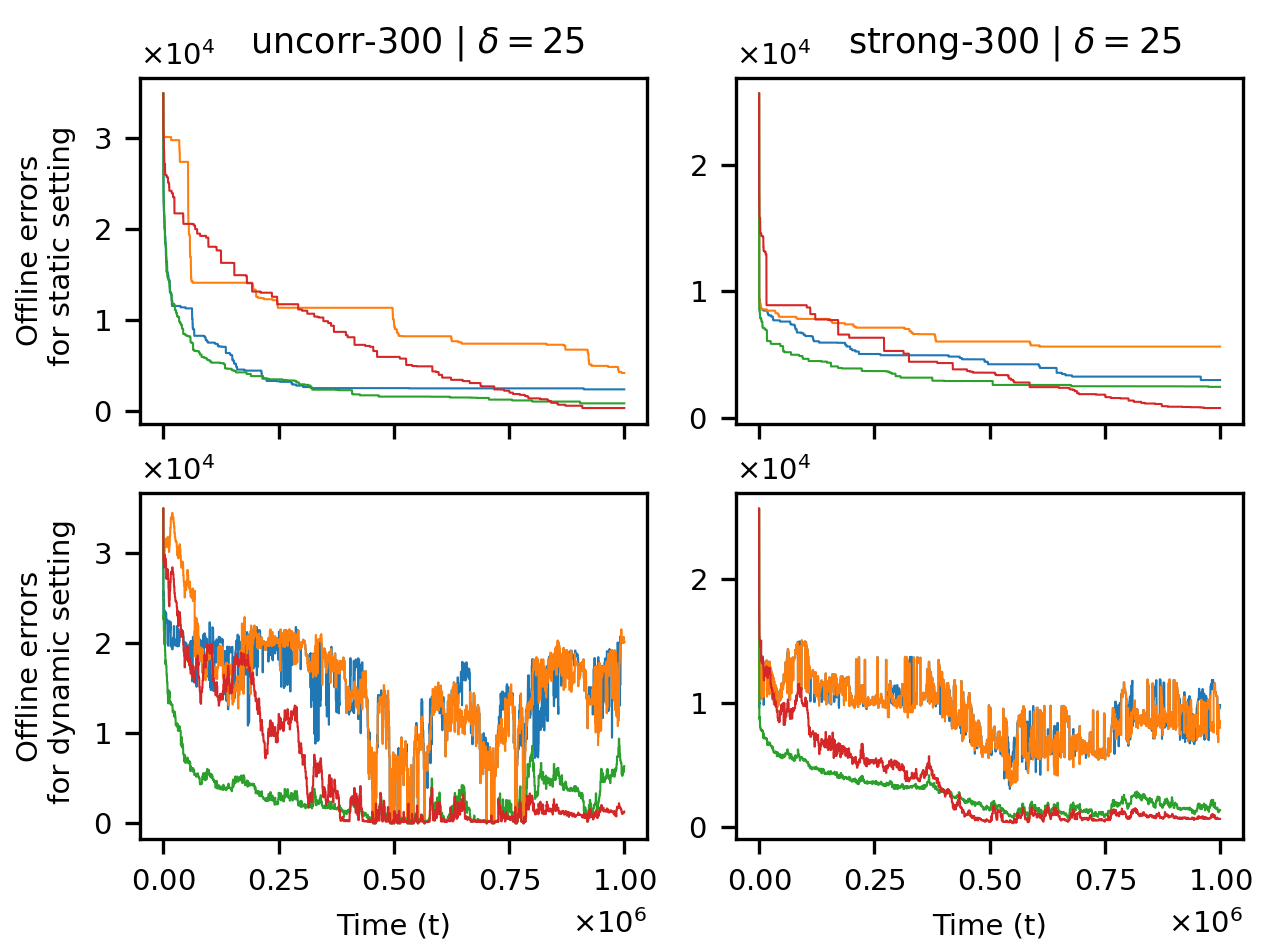}
    \end{tabular}
    \caption{Offline errors at each time step during optimisation of uncorr-300 and strong-300 for $\delta=25$ under static and dynamic ($\tau=1000,\gamma=1000$) settings.}
    \label{fig:error_vs_time}
\end{figure}
\begin{figure*}[!t]
    \centering
    \small
    \begin{tabular}{c}
         \includegraphics[scale=0.65]{new_tab/legend.png}\\
         \includegraphics[scale=0.75]{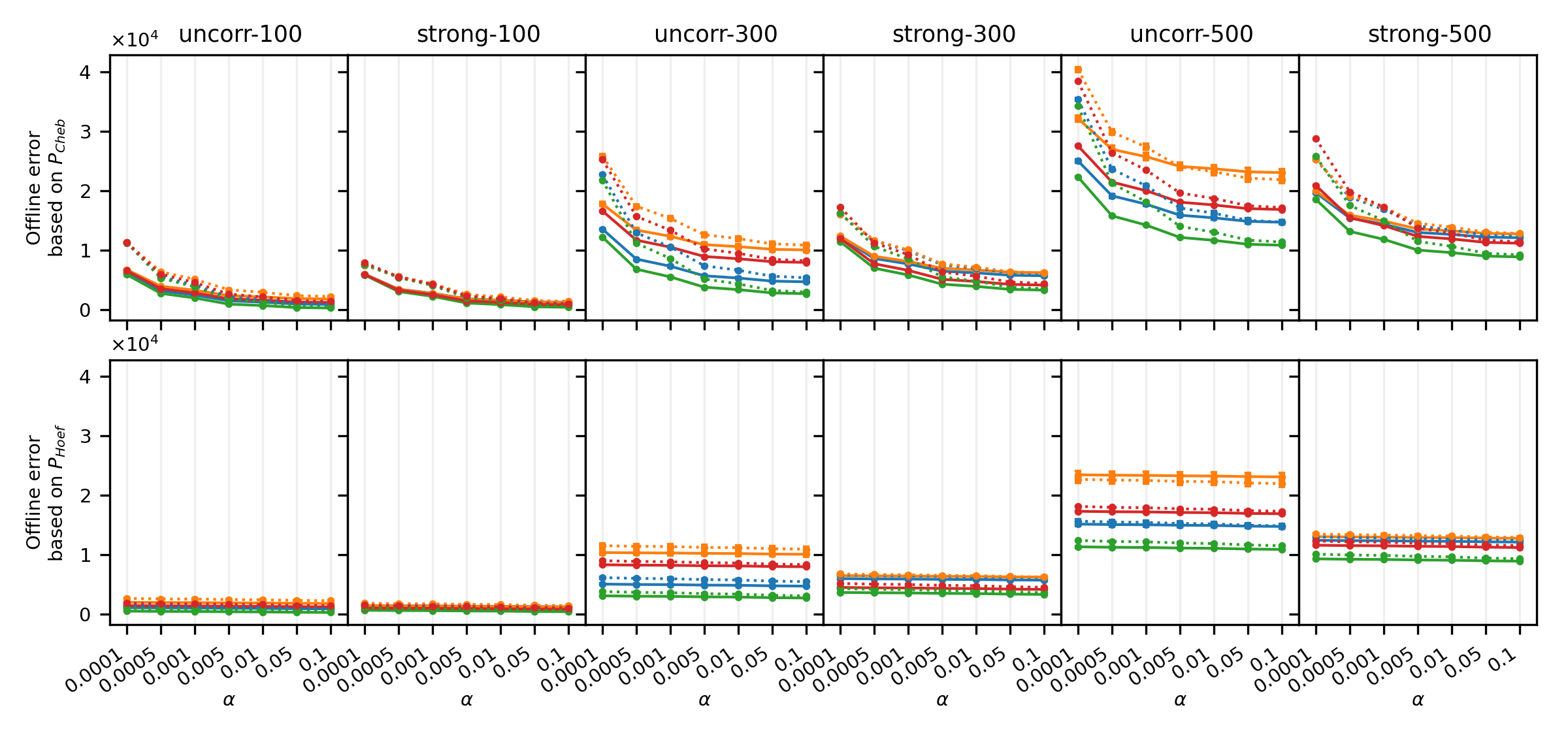}
    \end{tabular}
    \caption{Offline errors for static bound setting}
    \label{fig:static_err_results}
\end{figure*}

\section{Experimental Setup} \label{sec:setups}
Here, we present the details of the benchmark instances used for the experiments and design of experimental setups. 
We adopt the problem instances from the \cite{PISINGER20052271}. We consider two types of benchmarks based on the correlation between the profits and weights of elements: uncorrelated benchmarks and bounded and strongly correlated benchmarks with $n=100, 300$ and $500$ items. We identify these benchmarks as uncorr-100, uncorr-300, uncorr-500, strong-100, strong-300 and strong-500. 

First, we conduct experiments for the knapsack problem with a chance constraint on profits and a static constraint on weights. We consider that the weights of the elements are deterministic and use the values as given in the benchmarks. We define the profit $p_i$ of the element $i$ to be uniformly distributed as $p_i=[\mu_i-\delta, \mu_i+\delta]$ where $\mu_i$ is the expected profit of $i$ as given by the benchmark data. As the expected value of profits ($\mu_i$), we use the profits given in the benchmarks. We conduct the experiments under two stochastic settings,  $\delta=25, 50$. Next, we consider the knapsack problem with stochastic profits and a dynamic weight bound changing over time. We chose the initial weight bound $B_0$ of the problem instances as in the static weight bound scenario. Then, for every $\tau$ fitness evaluations, we select a new weight bound as $B_{t+1}=B_t+\gamma_t$, where $\gamma_t$ chosen uniformly at random from $[-\gamma,\gamma]$. We use the $\tau=1000, 2000$ and $\gamma=500, 1000$ in the experiments.

We consider GSEMO with standard parent selection, which is selecting a parent uniformly randomly from its population $P$ and also with sliding window-based parent selection. The original work that introduced sliding window selection for GSEMO \cite{neumann2023fast} used the sliding window length 1, which did not give us good results for our problems in the initial investigations. The target problems in the original work \cite{neumann2023fast} only consider problem settings where all weights are one or all weights are with the expected value of one. After preliminary experiments, we select the average weight of all elements  $len_{sw}$. Thus, the sliding window length ($len_{sw}$)  differs when running algorithms for different benchmarks. 

As the proposed fitness formulations optimise the problems independent of specific $\alpha$ values, we evaluate the outcome using the profit values for several $\alpha$ values between $0.1$ and $0.0001$. Here, we use the profit estimates $P_{Cheb}$ and $P_{Hoef}$ (see Equation \ref{eq:profit_cheby} and \ref{eq:profit_hoef}). Given a population of solutions $S$, the best solution $x^*$ with the confidence level $\alpha$ is identified using $P_{Cheb}$ and $P_{Hoef}$ as, $$\arg \max_{x \in S} P_{Cheb}(x,\alpha) \text{  and   } \arg \max_{x \in S} P_{Hoef}(x,\alpha)$$

In addition to the profits, we use a metric based on the error in each iteration specifically to evaluate the performance of evolutionary approaches. Offline error is a measure that has been successfully used in previous works on the optimisation of dynamic problems \cite{10.1007/978-3-319-99253-2_13, Hirad_ecai_20}. The offline error considers the distance of the best solution in each iteration to the deterministic optimum. Let $P_{opt}$ denote the deterministic optimum, the offline error for the confidence level $\alpha$ at time $t$ is given based on the two profit estimates as,
$$P_{opt} - \max_{x \in S} P_{Cheb}(x,\alpha) \text { and }P_{opt} - \max_{x \in S} P_{Hoef}(x,\alpha)$$

We use SCIP solver \cite{achterberg2009scip} to calculate $P_{opt}$ for each weight bound in dynamic problem instances. While the profit estimates give the quality of the final population, the offline errors indicate how the population evolves. This error measure is particularly beneficial to see how an evolutionary method deals with a moving constraint bound.
We present an analysis of results using both these measures in the following Section \ref{sec:results}.
\begin{figure*}[!t]
    \centering
    \small
    \begin{tabular}{c}
        \includegraphics[scale=0.65]{new_tab/legend.png}\\
        \includegraphics[scale=0.75]{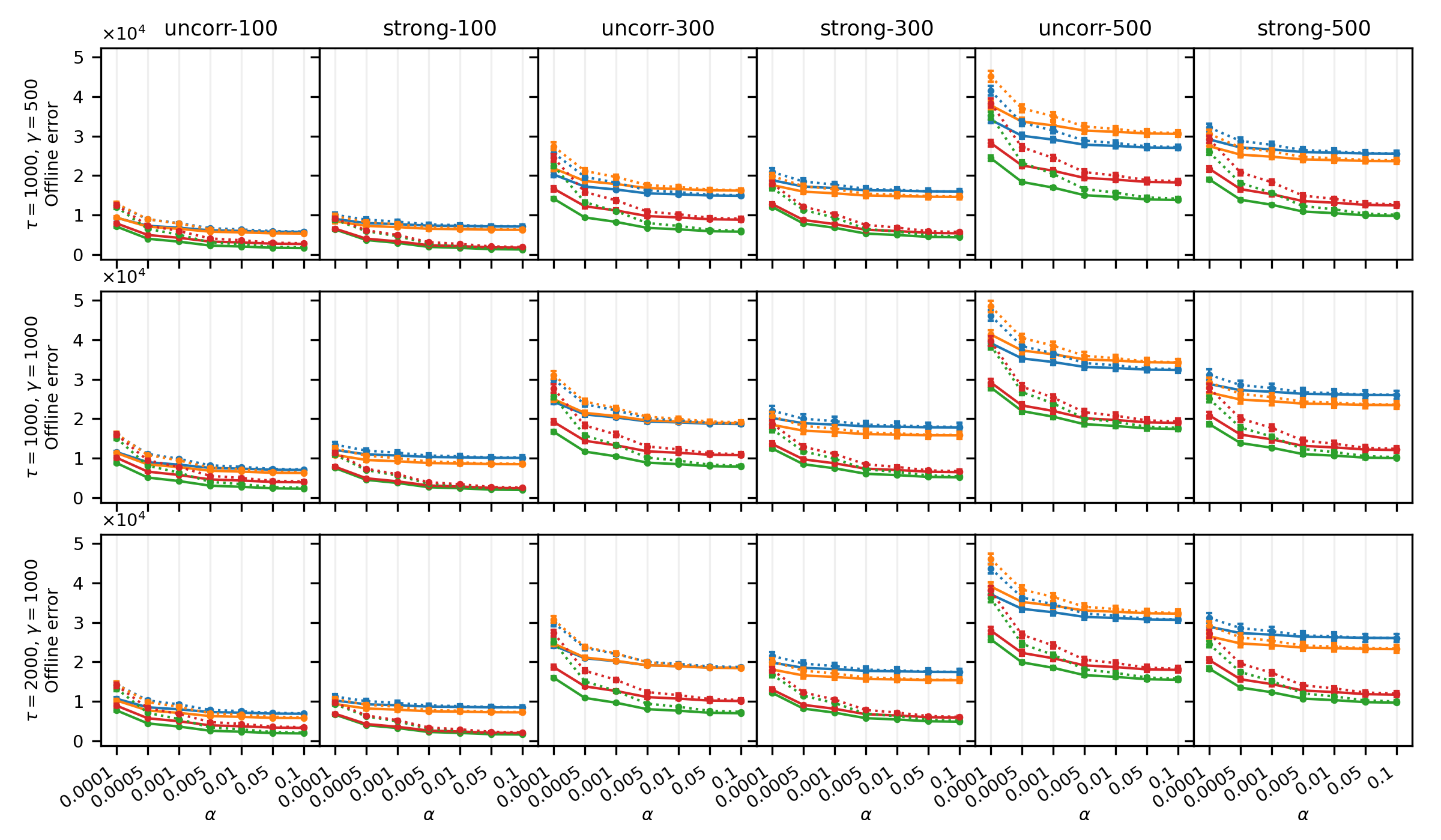}
    \end{tabular}
    \caption{Average offline error values based on $P_{Cheb}$ for dynamic settings}
    \label{fig:dyn_cheb}
\end{figure*}
\begin{figure*}[!t]
    \centering
    \small
    \begin{tabular}{c}
        \includegraphics[scale=0.65]{new_tab/legend.png} \\
        \includegraphics[scale=0.75]{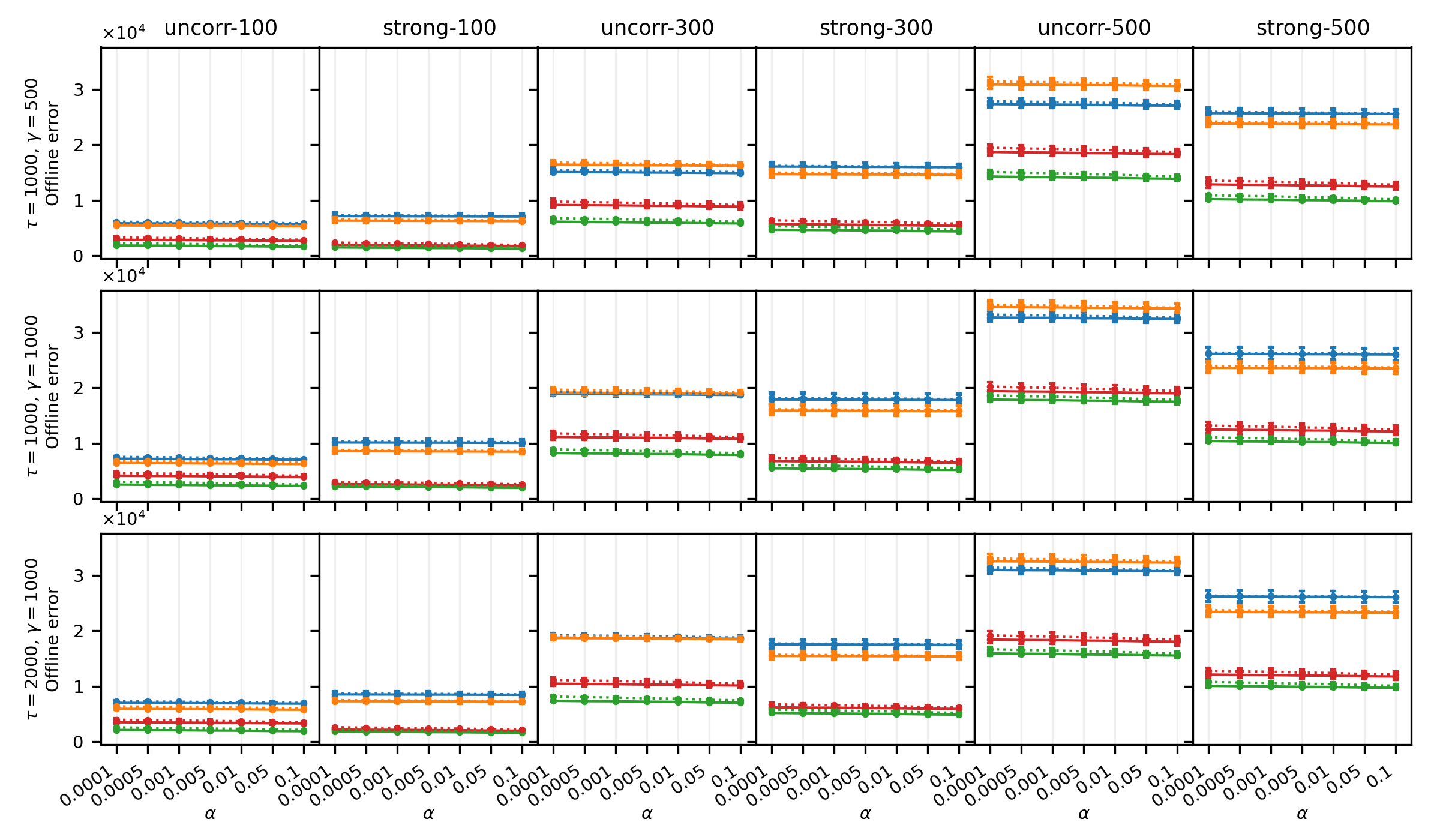}\\
    \end{tabular}
    
\caption{Average offline error values based on $P_{Hoef}$ for dynamic setting with $\tau=1000$ and $\gamma=1000$}
\label{fig:dyn_hoef}
\end{figure*}
\section{Experimental Results\label{sec:results}}
This section presents the profits and offline errors obtained for the target problem under static and dynamic weight bound. We use abbreviations to refer to the algorithmic approaches. We refer to GSEMO using the bi-objective and 3-objective formulations as GS-2D and GS-3D, respectively. We refer to the GSEMO that uses the sliding window selection with bi-objective and 3-objective formulations as SW-2D and SW-3D, respectively.

\begin{figure*}[!h]
    \centering
    \small
    \begin{tabular}{c}
        \includegraphics[scale=0.65]{new_tab/legend.png} \\
        \includegraphics[scale=0.75]{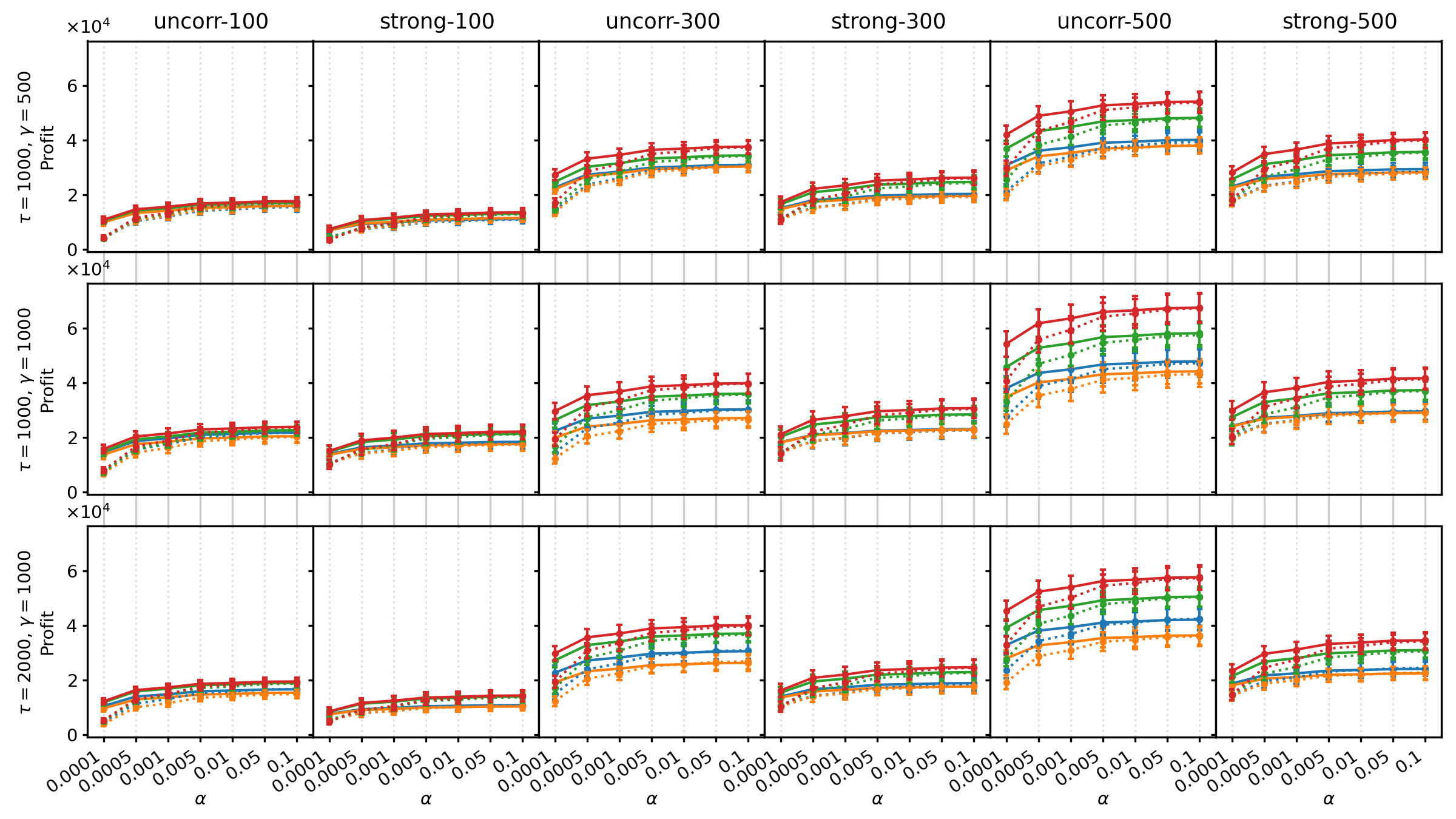}\\
    \end{tabular}
    
\caption{Best profits for the final dynamic state based on $P_{Cheb}$}
\label{fig:dyn_profit_cheb}
\end{figure*}

\begin{figure*}[!h]
    \centering
    \small
    \begin{tabular}{c}
        \includegraphics[scale=0.65]{new_tab/legend.png} \\
        \includegraphics[scale=0.75]{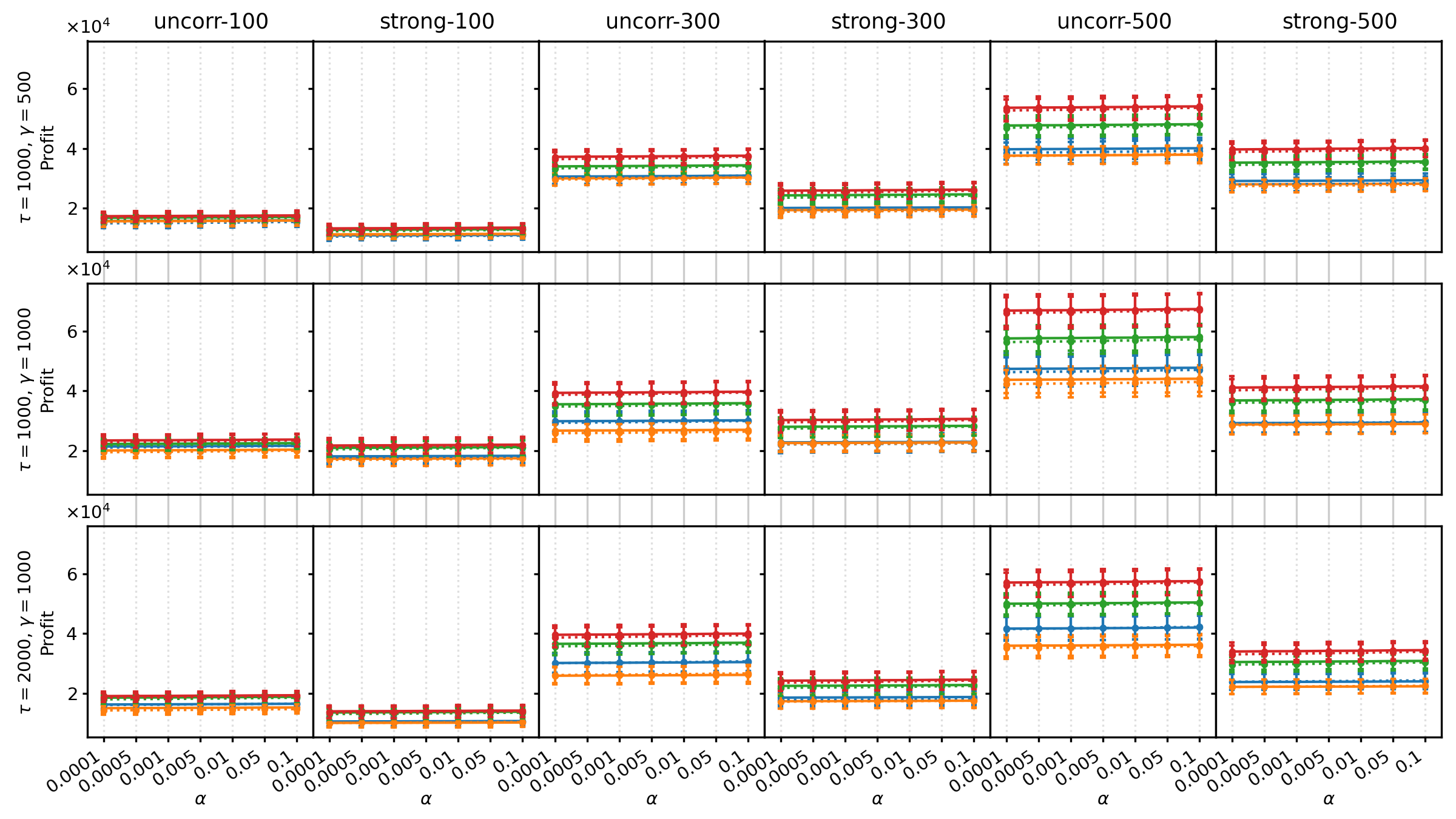}\\
    \end{tabular}
    
\caption{Best profits for the final dynamic state based on $P_{Hoef}$}
\label{fig:dyn_profit_hoef}
\end{figure*}

\subsection{Results for the knapsack problem with stochastic profits and static weight constraint \label{sec:results_static}}
This section discusses the results for the target problem under static weight bound value. Figure \ref{fig:static_results} visualises the profits given by the final population of each method for a range of $\alpha$ values. The $P_{Cheb}$ results show profits significantly increase as $\alpha$ increases while $P_{Hoef}$ values only increase slightly. The results from the four methods for small instances, uncorr-100 and strong-100, are similar. For other problem instances, we can see the outstanding results from 3-objective approaches over the bi-objective approaches. 

The bi-objective optimisation results for larger instances ($n=300, 500$) show that GS-2D gives better profit estimates than SW-2D. On the contrary, among the 3-objective approaches, SW-3D gives higher profit values than GS-3D. Here, as the problem size ($n$) increases, compared to GS-3D, SW-3D gives much better profit estimates. This observation for the 3-objective optimisation results is significant since the population size here is quite large compared to bi-objective methods. The integration of the sliding window technique has enabled the performance gain of GSEMO in 3-objective optimisation to improve further.

Next, we present how the offline errors change when optimising two selected problems in Figure \ref{fig:error_vs_time}. The first two plots consider a static bound setting and $\alpha=0.1$. As these plots depict, the SW-2D gives higher error values. We believe the reason is that in SW-2D, although selection considers the weight of the solution, the objectives do not represent the weight value of a solution. However, when using SW-3D, where there is an objective indicating the weight of the solutions, the sliding window technique gives significant results. However, the error values from SW-2D and SW-3D at the beginning of the optimisation are much higher than those of GS-2D and GS-3D. Therefore, towards the end of the optimisation, SW-3D gives lower errors and performs better than GS-3D. 

To summarise the results from the 30 algorithm runs for all experiment settings, we use the average offline errors as given in  Figure \ref{fig:static_err_results}. The offline errors decline as the $\alpha$ value increases. While the chance-constrained profits are closer to the deterministic optimum for higher $\alpha$, the average error values are low. Moreover, the offline errors based on $P_{Cheb}$ significantly decline over the increasing $\alpha$ while the errors based on $P_{Hoef}$ slightly decrease.

According to average offline errors, GSEMO with standard selection gives better (lower) error values than GSEMO with sliding window selection in both bi-objective and three-objective optimation. Although the average offline errors of SW-3D are lower, its difference from GS-3D indicates SW-3D performs well despite getting large error values at the initial steps of the optimisation (see Figure \ref{fig:error_vs_time}).

\subsection{Results for the knapsack problem with stochastic profits and dynamic weight constraint}
This subsection presents results for the dynamic and profit-chance-constrained knapsack problem. In these experiments, we only apply the modified GSEMO given in Algorithm \ref{alg:GSEMO_dyn} and fitness functions $f_{2D-dyn}$ and $f_{3D-dyn}$. However, we refer to the modified algorithm as \emph{GSEMO} for conciseness and use the abbreviations GS-2D and GS-3D.

We record the profits and offline error values for the dynamic and profit-chance-constrained knapsack problem. In this paper, we present the results for the dynamic problem using offline error values as they indicate how the optimisation techniques have handled the dynamic changes over time.
We present the average offline error values calculated based on $P_{Cheb}$ and $P_{Hoef}$ for dynamic settings in Figure \ref{fig:dyn_cheb} and \ref{fig:dyn_hoef}, respectively.

In all dynamic settings, three-objective approaches give much lower error values than bi-objective approaches. In bi-objective and three-objective optimisation, the average error values given by GSEMO when using standard selection are lower (better) than when using sliding window selection. As the last two plots in Figure\ref{fig:error_vs_time} show, even in the dynamic settings, the errors at the beginning of optimisation are higher for sliding window approaches than for standard selection approaches. However, towards the later iterations, offline errors are lower when using sliding window selection.

Therefore, we compare the error values in the dynamic settings against those in static settings, which gives us valuable insights into the performance of two selection methods used with GSEMO. Compared to the static settings of the weight bound (Figure \ref{fig:static_err_results}), in dynamic settings (see Figure \ref{fig:dyn_cheb} and \ref{fig:dyn_hoef}), the difference in the error values between the GS-2D and SW-2D as well as between GS-3D and SW-3D is much lower. Even though SW-2D and SW may give larger error values in the initial part of the optimisation compared to GS-2D and GS-3D, the average offline of GSEMO with sliding window errors is much closer to their counterparts of GSEMO with standard selection.

Finally, we present the best profits for different confidence levels given by each algorithm for the final dynamic state of the problems. The results concerning the $P_{Cheb}$ and $P_{Hoef}$ are given in Figure \ref{fig:dyn_profit_cheb} and \ref{fig:dyn_profit_hoef}. Since the sequence of dynamic change differs in each of the 30 runs, the final state of the dynamic bound may vary in different runs. Although it is not being the ideal indication of the algorithms' performance in dynamic settings, the patterns of these profit values are interesting. 

We can see the mean profit values of SW-3D are the highest. The GS-3D also give competitive profit estimates, while the bi-objective optimisation results are similar for both SW-2D and GS-2D. These plots further prove that, even though average offline errors are high, SW-3D performs the best towards the end of the optimisation.

\section{Conclusions}\label{sec:conclusions}

This study targets two variations of the knapsack problem with stochastic profits under static and dynamic weight limits. We introduce bi-objective and three objective fitness formulations for these problems. We explore the performance of GSEMO on these problems with its standard parent selection method and parent selection method based on a sliding window technique. Moreover, to tackle the dynamic weight bound, we modify the GSEMO algorithm by adding a \emph{backlog-like} population to store some infeasible solutions that can become useful in the subsequent dynamic change. 
Our experimental results significantly improve when the number of objectives increases from two to three. The sliding window technique for parent selection works well, specifically in three-objective optimisation to deal with huge population sizes. 

\section*{Acknowledgements}
This work has been supported with supercomputing resources provided by the Phoenix HPC service at the University of Adelaide.

\bibliographystyle{ACM-Reference-Format}
\bibliography{main.bbl}

\end{document}